%% file: acl2018.tex
\documentclass[11pt,a4paper]{article}
\usepackage[hyperref]{acl2018}
\usepackage{times}
\usepackage{amsfonts}
\usepackage{latexsym}
\usepackage{xcolor}
\usepackage{url}
\usepackage{wrapfig}
\usepackage{latexsym}
\usepackage{multirow}
\usepackage{mathtools}
\usepackage{calc}
\usepackage{amsmath, amsthm, amssymb, amsfonts}
\usepackage{url}
\usepackage{framed}
\usepackage{textcomp}
\usepackage{mathtools}

\aclfinalcopy 
\def\aclpaperid{595} 

\DeclareMathOperator{\LogSumExp}{LogSumExp}
\DeclareMathOperator{\csim}{csim}

\DeclarePairedDelimiter\floor{\lfloor}{\rfloor}

\newcommand\blfootnote[1]{%
  \begingroup
  \renewcommand\thefootnote{}\footnote{#1}%
  \addtocounter{footnote}{-1}%
  \endgroup
}

\title{Hierarchical Losses and New Resources for \\ Fine-grained Entity Typing and Linking}

\author{Shikhar Murty* \\
  UMass Amherst \\
  {\tt smurty@cs.umass.edu} \\\And
  Patrick Verga* \\
  UMass Amherst \\
  {\tt pat@cs.umass.edu} \\\And
  Luke Vilnis \\
  UMass Amherst \\
  {\tt luke@cs.umass.edu}  \\\AND
  Irena Radovanovic \\
  Chan Zuckerberg Initiative \\
  {\tt iradovanovic@chanzuckerberg.com}  \\\And
  Andrew McCallum \\
  UMass Amherst \\
  {\tt mccallum@cs.umass.edu}
  }

\date{}

\begin{document}
\maketitle

\begin{abstract}
Extraction from raw text to a knowledge base of entities and fine-grained types is often cast as prediction into a flat set of entity and type labels, neglecting the rich hierarchies over types and entities contained in curated ontologies. Previous attempts to incorporate hierarchical structure have yielded little benefit and are restricted to shallow ontologies. This paper presents new methods using real and complex bilinear mappings for integrating hierarchical information, yielding substantial improvement over flat predictions in entity linking and fine-grained entity typing, and achieving new state-of-the-art results for end-to-end models on the benchmark FIGER dataset. We also present two new human-annotated datasets containing wide and deep hierarchies which we will release to the community to encourage further research in this direction: \emph{MedMentions}, a collection of PubMed abstracts in which 246k mentions have been mapped to the massive UMLS ontology; and \emph{TypeNet}, which aligns Freebase types with the WordNet hierarchy to obtain nearly 2k entity types. In experiments on all three datasets we show substantial gains from hierarchy-aware training. 
\blfootnote{*equal contribution}
\blfootnote{Data and code for experiments: \url{https://github.com/MurtyShikhar/Hierarchical-Typing}}
\end{abstract}

\input{intro}

\input{data_sets}

\input{model}

\input{experiments}

\input{related_work}

\section{Conclusion}
We demonstrate that explicitly incorporating and modeling hierarchical information leads to increased performance in experiments on entity typing and linking across three challenging datasets. Additionally, we introduce two new human-annotated datasets: MedMentions, a corpus of 246k mentions from PubMed abstracts linked to the UMLS knowledge base, and TypeNet, a new hierarchical fine-grained entity typeset an order of magnitude larger and deeper than previous datasets.

While this work already demonstrates considerable improvement over non-hierarchical modeling, future work will explore techniques such as Box embeddings \cite{acl18-vilnis} and Poincar{\'e} embeddings \cite{nickel2017poincar} to represent the hierarchical embedding space, as well as methods to improve recall in the candidate generation process for entity linking. Most of all, we are excited to see new techniques from the NLP community using the resources we have presented.

\section{Acknowledgements}
We thank Nicholas Monath, Haw-Shiuan Chang and Emma Strubell for helpful comments on early drafts of the paper. Creation of the MedMentions corpus is supported and managed by the Meta team at the Chan Zuckerberg Initiative. A pre-release of the dataset is available at \url{http://github.com/chanzuckerberg/MedMentions}. This work was supported in part by the Center for Intelligent Information Retrieval and the Center for Data Science, in part by the Chan Zuckerberg Initiative under the project Scientific Knowledge Base Construction., and in part by the National Science Foundation under Grant No. IIS-1514053. Any opinions, findings and conclusions or recommendations expressed in this material are those of the authors and do not necessarily reflect those of the sponsor.

\bibliography{acl2018}
\bibliographystyle{acl_natbib}

\clearpage
\appendix
\input{appendix}

\end{document}

%% file: intro.tex
\section{Introduction}

Identifying and understanding entities is a central component in knowledge base construction \cite{roth2015building} and essential for enhancing downstream tasks such as relation extraction \cite{yaghoobzadeh-adel-schutze:2017:EACLlong}, question answering \cite{das-EtAl:2017:Short, welbl2017constructing} and search \cite{dalton2014entity}. This has led to considerable research in automatically identifying entities in text, predicting their types, and linking them to existing structured knowledge sources.

Current state-of-the-art models encode a textual mention with a neural network and classify the mention as being an instance of a fine grained type or entity in a knowledge base.  Although in many cases the types and their entities are arranged in a hierarchical ontology, most approaches ignore this structure, and previous attempts to incorporate hierarchical information yielded little improvement in performance \citep{shimaoka-EtAl:2017:EACLlong}. 
Additionally, existing benchmark entity typing datasets only consider small label sets arranged in very shallow hierarchies. For example, FIGER \cite{ling2012fine}, the \emph{de facto} standard fine grained entity type dataset, contains only 113 types in a hierarchy only two levels deep. 

In this paper we investigate models that explicitly integrate hierarchical information into the embedding space of entities and types, using a hierarchy-aware loss on top of a deep neural network classifier over textual mentions. By using this additional information, we learn a richer, more robust representation, gaining statistical efficiency when predicting similar concepts and aiding the classification of rarer types. We first validate our methods on the narrow, shallow type system of FIGER, 
out-performing state-of-the-art methods not incorporating hand-crafted features and matching those that do.

To evaluate on richer datasets and stimulate further research into hierarchical entity/typing prediction with larger and deeper ontologies, we introduce two new human annotated datasets. The first is {\it MedMentions}, a collection of PubMed abstracts in which 246k concept mentions have been annotated with links to the Unified Medical Language System (UMLS) ontology \cite{bodenreider2004unified}, an order of magnitude more annotations than comparable datasets. UMLS contains over 3.5 million concepts in a hierarchy having average depth 14.4. Interestingly, UMLS does not distinguish between types and entities (an approach we heartily endorse), and the technical details of linking to such a massive ontology lead us to refer to our MedMentions experiments as entity linking. Second, we present {\it TypeNet}, a curated mapping from the Freebase type system into the WordNet hierarchy. TypeNet contains over 1900 types with an average depth of 7.8.

In experimental results, we show improvements with a hierarchically-aware training loss on each of the three datasets. In entity-linking MedMentions to UMLS, we observe a 6\% relative increase in accuracy over the base model. In experiments on entity-typing from Wikipedia into TypeNet, we show that incorporating the hierarchy of types and including a hierarchical loss provides a dramatic 29\% relative increase in MAP. Our models even provide benefits for shallow hierarchies allowing us to match the state-of-art results of \citet{shimaoka-EtAl:2017:EACLlong} on the FIGER (GOLD) dataset without requiring hand-crafted features.

We will publicly release the TypeNet and MedMentions datasets to the community to encourage further research in truly fine-grained, hierarchical entity-typing and linking.

%% file: data_sets.tex
\section{New Corpora and Ontologies}

\subsection{MedMentions \label{sec:pubmed_umls}}
Over the years researchers have constructed many large knowledge bases in the biomedical domain \cite{apweiler2004uniprot, davis2008comparative, chatr2017biogrid}. Many of these knowledge bases are specific to a particular sub-domain encompassing a few particular types such as genes and diseases \cite{pinero2017disgenet}. 

UMLS \cite{bodenreider2004unified} is particularly comprehensive, containing over 3.5 million concepts (UMLS does not distinguish between entities and types) defining their relationships and a  curated hierarchical ontology. For example \textit{LETM1 Protein} {\sc is-a} \textit{Calcium Binding Protein} {\sc is-a} \textit{Binding Protein} {\sc is-a} \textit{Protein} {\sc is-a} \textit{Genome Encoded Entity}. This fact makes UMLS particularly well suited for methods explicitly exploiting hierarchical structure.

Accurately linking textual biological entity mentions to an existing knowledge base is extremely important but few richly annotated resources are available. Even when resources do exist, they often contain no more than a few thousand annotated entity mentions which is insufficient for training state-of-the-art neural network entity linkers. State-of-the-art methods must instead rely on string matching between entity mentions and canonical entity names \cite{leaman2013dnorm,wei2015gnormplus,leaman2016taggerone}. 
To address this, we constructed MedMentions, a new, large dataset identifying and linking entity mentions in PubMed abstracts to specific UMLS concepts. Professional annotators exhaustively annotated UMLS entity mentions from 3704 PubMed abstracts, resulting in 246,000 linked mention spans. The average depth in the hierarchy of a concept from our annotated set is 14.4 and the maximum depth is 43. 

MedMentions\footnote{\protect{\url{http://github.com/chanzuckerberg/MedMentions}}} contains an order of magnitude more annotations than similar biological entity linking PubMed datasets \citep{dougan2014ncbi, wei2015gnormplus, li2016biocreative}. Additionally, these datasets contain annotations for only one or two entity types (genes or chemicals and disease etc.). MedMentions instead contains annotations for a wide diversity of entities linking to UMLS. Statistics for several other datasets are in Table \ref{tab:medmentions_compare_stats} and further statistics are in \ref{tab:medmentions_stats}.

\begin{table}[htp]{
\small
\centering
\begin{tabular}{|l|c|c|} \hline
    Dataset & mentions & unique entities \\ \hline \hline
    MedMentions & 246,144 & 25,507 \\
    \hline
    BCV-CDR & 28,797 & 2,356\\ 
	NCBI Disease & 6,892 & 753 \\
    BCII-GN Train & 6,252 & 1,411 \\
    NLM Citation GIA & 1,205 & 310\\
\hline
  \end{tabular}
  \caption{Statistics from various biological entity linking data sets from scientific articles. NCBI Disease \citep{dougan2014ncbi} focuses exclusively on disease entities. BCV-CDR \citep{li2016biocreative} contains both chemicals and diseases. BCII-GN and NLM \citep{wei2015gnormplus} both contain genes. \label{tab:medmentions_compare_stats}}
}
\end{table}

\begin{table}[htp]
\centering
{	
\small
\begin{tabular}{|l|r|r|r|} \hline
		Statistic & Train & Dev & Test \\ \hline \hline
		\#Abstracts &  2,964 & 370 & 370 \\
		\#Sentences &  28,457 & 3,497 & 3,268 \\
		\#Mentions & 199,977 & 24,026 & 22,141 \\
		\#Entities &  22,416 & 5,934 & 5,521 \\ \hline
	\end{tabular}
	\caption{MedMentions statistics.  \label{tab:medmentions_stats}}
}
\end{table}

\subsection{TypeNet}


TypeNet\footnote{\protect{\url{https://github.com/iesl/TypeNet}}} is a new dataset of hierarchical entity types for extremely fine-grained entity typing. TypeNet was created by manually aligning Freebase types \cite{bollacker2008freebase} to noun synsets from the WordNet hierarchy \citep{fellbaum1998wordnet}, naturally producing a hierarchical type set.

To construct TypeNet, we first consider all Freebase types that were linked to more than 20 entities. This is done to eliminate types that are either very specific or very rare. We also remove all Freebase API types, e.g. the {[{/freebase}, {/dataworld}, {/schema}, {/atom}, {/scheme}, and {/topics}]} domains. 

For each remaining Freebase type, we generate a list of candidate WordNet synsets through a substring match. An expert annotator then attempted to map the Freebase type to one or more synsets in the candidate list with a \emph{parent-of}, \emph{child-of} or \emph{equivalence} link by comparing the definitions of each synset with example entities of the Freebase type. If no match was found, the annotator manually formulated queries for the online WordNet API until an appropriate synset was found. See Table \ref{table:annotate} for an example annotation.

Two expert annotators independently aligned each Freebase type before meeting to resolve any conflicts. The annotators were conservative with assigning equivalence links resulting in a greater number of \emph{child-of} links. The final dataset contained 13 \emph{parent-of}, 727 \emph{child-of}, and 380 \emph{equivalence} links. Note that some Freebase types have multiple \emph{child-of} links to WordNet, making TypeNet, like WordNet, a directed acyclic graph. We then took the union of each of our annotated Freebase types, the synset that they linked to, and any ancestors of that synset. 

\begin{table}[htp]{
\small
\centering
\begin{tabular}{|l|r|r|c|} \hline
    Typeset & Count & Depth & Gold KB links \\ \hline \hline
    CoNLL-YAGO &  4  & 1 & Yes \\
    OntoNotes 5.0 &  19  & 1 & No \\
    \citet{DBLP:journals/corr/GillickLGKH14} & 88 & 3 & Yes \\
    Figer & 112 & 2 & Yes \\
    Hyena & 505 & 9 & No\\
    Freebase & 2k & 2 & Yes \\
	WordNet & 16k & 14 & No \\ 
    TypeNet* & 1,941 & 14 & Yes \\ \hline
  \end{tabular}
  \caption{Statistics from various type sets. TypeNet is the largest type hierarchy with a gold mapping to KB entities. *The entire WordNet could be added to TypeNet increasing the total size to 17k types.  \label{data:stats}}
}
\end{table}


We also added an additional set of 614 \emph{FB} $\rightarrow$ \emph{FB} links \ref{data:typenet_stats}. This was done by computing conditional probabilities of Freebase types given other Freebase types from a collection of 5 million randomly chosen Freebase entities. The conditional probability P($t_2 \mid t_1$) of a Freebase type $t_2$ given another Freebase type $t_1$ was calculated as $\frac{\#(t_1, t_2)}{\#t_1}$. Links with a conditional probability less than or equal to 0.7 were discarded. The remaining links were manually verified by an expert annotator and valid links were added to the final dataset, preserving acyclicity. 

\begin{table}[htp]{
\small
\centering
\begin{tabular}{|l|l|}
\hline
    Freebase Types & 1081 \\
    WordNet Synsets & 860 \\
    \hline
    child-of links & 727 \\
    equivalence links & 380 \\
    parent-of links & 13 \\
    Freebase-Freebase links & 614 \\
    \hline
  \end{tabular}
  \caption{Stats for the final TypeNet dataset. child-of, parent-of, and equivalence links are from Freebase types $\rightarrow$ WordNet synsets. \label{data:typenet_stats}}
}
\end{table}

%% file: model.tex
\section{Model}
\subsection{Background: Entity Typing and Linking}
We define a textual mention $m$ as a sentence with an identified entity. The goal is then to classify $m$ with one or more labels. For example, we could take the sentence $m=$ ``\textit{Barack Obama is the President of the United States.}'' with the identified entity string \textbf{Barack Obama}. In the task of \textit{entity linking}, we want to map $m$ to a specific entity in a knowledge base such as ``m/02mjmr'' in Freebase. In \textit{mention-level typing}, we label $m$ with one or more types from our type system $T$ such as $t^m$ = \{president, leader, politician\} \cite{ling2012fine,DBLP:journals/corr/GillickLGKH14,shimaoka-EtAl:2017:EACLlong}.  In \textit{entity-level typing}, we instead consider a bag of mentions $B_e$ which are all linked to the same entity. We label $B_e$ with $t^e$, the set of all types expressed in all $m \in B_e$ \cite{yao2013universal, neelakantan-chang:2015:NAACL-HLT, verga-neelakantan-mccallum:2017:EACLlong,yaghoobzadeh2017corpus}. 

\subsection{Mention Encoder \label{sec:encoder}}
Our model converts each mention $m$ to a $d$ dimensional vector. This vector is used to classify the type or entity of the mention. The basic model depicted in Figure \ref{fig:encoder} concatenates the averaged word embeddings of the mention string with the output of a convolutional neural network (CNN). The word embeddings of the mention string capture global, context independent semantics while the CNN encodes a context dependent representation. 

\subsubsection{Token Representation} 
Each sentence is made up of $s$ tokens which are mapped to $d_w$ dimensional word embeddings. Because sentences may contain mentions of more than one entity, we explicitly encode a distinguished mention in the text using position embeddings which have been shown to be useful in state of the art relation extraction models \citep{DBLP:conf/acl/SantosXZ15,lin2016neural} and machine translation \citep{DBLP:conf/nips/VaswaniSPUJGKP17}.
Each word embedding is concatenated with a $d_p$ dimensional learned position embedding encoding the token's relative distance to the target entity. Each token within the distinguished mention span has position 0, tokens to the left have a negative distance from $[-s, 0)$, and tokens to the right of the mention span have a positive distance from $(0,s]$. We denote the final sequence of token representations as $M$.

\subsubsection{Sentence Representation}
\begin{figure}[t!]
    \centering
    \includegraphics[width=0.35\textwidth, height=0.35\textwidth, scale=0.4]{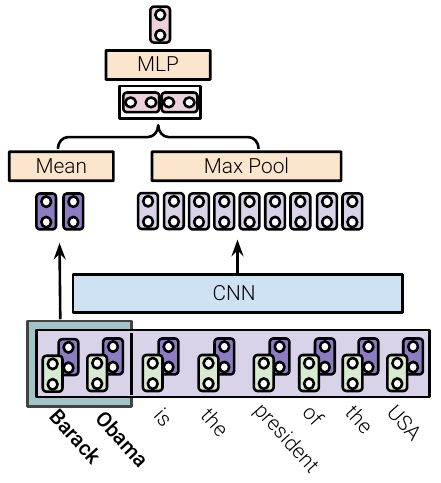}
    \caption{Sentence encoder for all our models. The input to the CNN consists of the concatenation of position embeddings with word embeddings. The output of the CNN is concatenated with the mean of mention surface form embeddings, and then passed through a 2 layer MLP. \label{fig:encoder}}
\end{figure}

The embedded sequence $M$ is then fed into our context encoder. Our context encoder is a single layer CNN followed by a $\tanh$ non-linearity to produce $C$. The outputs are max pooled across time to get a final context embedding, ${m}_{\mathrm{CNN}}$.
\begin{align*}
&{c_i} = \tanh ({b} + \sum_{j=0}^{w} W[j]M[i-\floor{\frac{w}{2}}+j]) \\
&{m}_{\text{CNN}} = \max_{0 \le i \le n-w+1} c_i
\end{align*}
Each $W[j] \in \mathbb{R}^{d \times d} $ is a CNN filter, the bias ${b} \in \mathbb{R}^d$, $M[i] \in \mathbb{R}^d$ is a token representation, and the max is taken pointwise. In all of our experiments we set $w=5$.

In addition to the contextually encoded mention, we create a global mention encoding, ${m}_{\text{G}}$, by averaging the word embeddings of the tokens within the mention span. 

The final mention representation $m_\text{F}$ is constructed by concatenating ${m}_{\text{CNN}}$ and ${m}_{\text{G}}$ and applying a two layer feed-forward network with $\tanh$ non-linearity (see Figure \ref{fig:encoder}):
\begin{align*}
m_\text{F} = W_2\tanh(W_1 \begin{bmatrix}{m}_{\text{SFM}} \\ {m}_{\text{CNN}}\end{bmatrix} + b_1) + b_2
\end{align*}

\section{Training}
\subsection{Mention-Level Typing \label{sec:mention_typing}}
Mention level entity typing is treated as multi-label prediction. Given the sentence vector $m_\text{F}$, we compute a score for each type in typeset $T$ as:
\begin{align*}
y_j  =  \mathbf{t_j}^\top m_\text{F}
\end{align*}
where $\mathbf{t_j}$ is the embedding for the j\textsuperscript{th} type in $T$ and $y_j$ is its corresponding score. The mention is labeled with $t^{m}$, a binary vector of all types where $t^m_j=1$ if the j\textsuperscript{th} type is in the set of gold types for $m$ and 0 otherwise. We optimize a multi-label binary cross entropy objective:
\begin{align*}
&\mathcal{L}_\text{type}(m) = - \sum_j t^m_j \log y_j + (1-t^m_j) \log (1 - y_j)
\end{align*}

\subsection{Entity-Level Typing \label{sec:entity_level_typing}}
In the absence of mention-level annotations, we instead must rely on distant supervision \cite{distant_supervision} to noisily label all mentions of entity $e$ with all types belonging to $e$. This procedure inevitably leads to noise as not all mentions of an entity express each of its known types. To alleviate this noise, we use multi-instance multi-label learning (MIML) \cite{surdeanu2012multi} which operates over \emph{bags} rather than mentions. A bag of mentions $B_e = \{{m^1}, {m^2}, \ldots , {m^n}\} $ is the set of all mentions belonging to entity $e$. The bag is labeled with $t^{e}$, a binary vector of all types where $t^e_j=1$ if the j\textsuperscript{th} type is in the set of gold types for $e$ and 0 otherwise.

For every entity, we subsample $k$ mentions from its bag of mentions. Each mention is then encoded independently using the model described in Section \ref{sec:encoder} resulting in a bag of vectors. Each of the $k$ sentence vectors $m_\text{F}^i$ is used to compute a score for each type in $t^e$: 
\begin{align*}
y^{i}_j  =  \mathbf{t_j}^\top m_\text{F}^i
\end{align*}
where $\mathbf{t_j}$ is the embedding for the j\textsuperscript{th} type in $t^e$ and $y^i$ is a vector of logits corresponding to the i\textsuperscript{th} mention. The final bag predictions are obtained using element-wise $\LogSumExp$ pooling across the $k$ logit vectors in the bag to produce entity level logits $y$:
\begin{align*}
&{y} =\log \sum_{i} {\exp({y^{i}} )}
\end{align*}
We use these final bag level predictions to optimize a multi-label binary cross entropy objective:
\begin{align*}
&\mathcal{L}_\text{type}(B_e) = - \sum_j t^e_j \log y_j + (1-t^e_j) \log (1 - y_j)
\end{align*}

\subsection{Entity Linking \label{sec:model_linking}}
Entity linking is similar to mention-level entity typing with a single correct class per mention. Because the set of possible entities is in the millions, linking models typically integrate an alias table mapping entity mentions to a set of possible candidate entities. Given a large corpus of entity linked data, one can compute conditional probabilities from mention strings to entities \cite{spitkovsky2012cross}. In many scenarios this data is unavailable. However, knowledge bases such as UMLS contain a canonical string name for each of its curated entities. State-of-the-art biological entity linking systems tend to operate on various string edit metrics between the entity mention string and the set of canonical entity strings in the existing structured knowledge base \cite{leaman2013dnorm,wei2015gnormplus}. 

For each mention in our dataset, we generate 100 candidate entities $e_c = (e_1, e_2, \ldots , e_{100})$ each with an associated string similarity score $\csim$. See Appendix \ref{sec:candidate_details} for more details on candidate generation. We generate the sentence representation $m_\text{F}$ using our encoder and compute a similarity score between $m_\text{F}$ and the learned embedding $e$ of each of the candidate entities. This score and string cosine similarity $\csim$ are combined via a learned linear combination to generate our final score. The final prediction at test time $\hat{e}$ is the maximally similar entity to the mention.
\begin{align*}
&\phi (m, e) = \alpha ~e^\top m_\text{F} + \beta \csim(m,e) \\
&\hat{e} = \underset{e \in e_c}{\mathrm{argmax}}\  \phi(m, e)
\end{align*}
We optimize this model by multinomial cross entropy over the set of candidate entities and correct entity $e$.
\begin{align*}
\mathcal{L}_\text{link}(m,e_c) =  -~ \phi (m, e) +   \log {\sum_{e' \in e_c} {\exp{\phi (m, e')}}}
\end{align*}

\section{Encoding Hierarchies \label{sec:structure_loss}}

Both entity typing and entity linking treat the label space as prediction into a flat set. To explicitly incorporate the structure between types/entities into our training, we add an additional loss. We consider two methods for modeling the hierarchy of the embedding space: real and complex bilinear maps, which are two of the state-of-the-art knowledge graph embedding models. 

\subsection{Hierarchical Structure Models}
\noindent \textbf{Bilinear}: Our standard bilinear model scores a hypernym link between $(c_1, c_2)$ as:
\begin{align*}
s(c_1, c_2)  =  {c_1}^\top A {c_2}
\end{align*}
 where A $\in \mathbb{R}^{d \times d}$ is a learned real-valued non-diagonal matrix and $c_1$ is the child of $c_2$ in the hierarchy. This model is equivalent to RESCAL \cite{DBLP:conf/icml/NickelTK11} with a single {\sc is-a} relation type. The type embeddings are the same whether used on the left or right side of the relation. We merge this with the base model by using the parameter $A$ as an additional map before type/entity scoring.

\noindent \textbf{Complex Bilinear}: We also experiment with a complex bilinear map based on the ComplEx model \cite{trouillon2016complex}, which was shown to have strong performance predicting the hypernym relation in WordNet, suggesting suitability for asymmetric, transitive relations such as those in our type hierarchy.
ComplEx uses complex valued vectors for types, and diagonal complex matrices for relations, using Hermitian inner products (taking the complex conjugate of the second argument, equivalent to treating the right-hand-side type embedding to be the complex conjugate of the left hand side), and finally taking the real part of the score\footnote{This step makes the scoring function technically not bilinear, as it commutes with addition but not complex multiplication, but we term it \emph{bilinear} for ease of exposition.}. The score of a hypernym link between $(c_1, c_2)$ in the ComplEx model is defined as:
\begin{align*}
s(c_1, c_2) &= \text{Re}(<{c_1}, r_{\text{{\sc Is-A}}} , {c_2} > )  \\
						 &= \text{Re} (\sum_{k} c_{1k} r_{k} \bar{c}_{2k}) \\
						 &= \langle \text{Re}(c_{1}), \text{Re}(r_{\text{{\sc Is-A}}}) , \text{Re}(c_{2})\rangle  \\
                         &+ \langle \text{Re}(c_{1}), \text{Im}(r_{\text{{\sc Is-A}}}) , \text{Im}(c_{2})\rangle \\
                         &+ \langle \text{Im}(c_{1}), \text{Re}(r_{\text{{\sc Is-A}}}) , \text{Im}(c_{2})\rangle  \\
                         &- \langle \text{Im}(c_{1}), \text{Im}(r_{\text{{\sc Is-A}}}) , \text{Re}(c_{2})\rangle  \\
\end{align*}  
where ${c_1}$, ${c_2}$ and $r_{\text{{\sc Is-A}}}$ are complex valued vectors representing $c_1$, $c_2$ and the {\sc is-a} relation respectively. $\text{Re}(z)$ represents the real component of $z$ and $\text{Im}(z)$ is the imaginary component. As noted in \citet{trouillon2016complex}, the above function is antisymmetric when ${r_{\text{{\sc is-a}}}}$ is purely imaginary.

Since entity/type embeddings are complex vectors, in order to combine it with our base model, we also need to represent mentions with complex vectors for scoring. To do this, we pass the output of the mention encoder through two different affine transformations to generate a real and imaginary component:
\begin{align*}
\text{Re}({{m_\text{F}}} ) &= W_\text{real} {m_\text{F}}  + b_\text{real} \\ 
\text{Im}({{m_\text{F}}}) &= W_\text{img} {m_\text{F}}  + b_\text{img}
\end{align*}
where $m_\text{F}$ is the output of the mention encoder, and $W_\text{real}$, $W_\text{img}$ $\in \mathbb{R}^{d \times d} $   and $b_\text{real}$, $b_\text{img}$ $\in \mathbb{R}^{d}$ .

\subsection{Training with Hierarchies}
Learning a hierarchy is analogous to learning embeddings for nodes of a knowledge graph with a single hypernym/{\sc is-a} relation. To train these embeddings, we sample $(c_{1}, c_{2})$ pairs, where each pair is a positive link in our hierarchy. For each positive link, we sample a set $N$ of $n$ negative links. We encourage the model to output high scores for positive links, and low scores for negative links via a binary cross entropy (BCE) loss:
\begin{align*}
\mathcal{L}_{\text{struct}} &=  -\log \sigma(s(c_{1i},  c_{2i})) \\&~~+ \sum_{N}\log(1 - \sigma(s(c_{1i} , c'_{2i}))) \\
\mathcal{L} &= \mathcal{L}_{\text{type/link}} + \gamma \mathcal{L}_{\text{struct}}
\end{align*}
where $s(c_1, c_2)$ is the score of a link $(c_1, c_2)$, and $\sigma(\cdot)$ is the logistic sigmoid. The weighting parameter $\gamma$ is $\in$ \{0.1, 0.5, 0.8, 1, 2.0, 4.0\}. The final loss function that we optimize is $\mathcal{L}$.

%% file: experiments.tex
\section{Experiments}
We perform three sets of experiments: mention-level entity typing on the benchmark dataset FIGER, entity-level typing using Wikipedia and TypeNet, and entity linking using MedMentions.

\subsection{Models}
\noindent\emph{\textbf{CNN}}: Each mention is encoded using the model described in Section \ref{sec:encoder}. The resulting embedding is used for classification into a flat set labels. Specific implementation details can be found in Appendix \ref{sec:model_implementation}.

\noindent\emph{\textbf{CNN+Complex}}: The CNN+Complex model is equivalent to the CNN model but uses complex embeddings and Hermitian dot products.

\noindent\emph{\textbf{Transitive}}: This model does not add an additional hierarchical loss to the training objective (unless otherwise stated). We add additional labels to each entity corresponding to the transitive closure, or the union of all ancestors of its known types. This provides a rich additional learning signal that greatly improves classification of  specific types.

\noindent\emph{\textbf{Hierarchy}}: These models add an explicit hierarchical loss to the training objective, as described in Section~\ref{sec:structure_loss}, using either complex or real-valued bilinear mappings, and the associated parameter sharing.

\subsection{Mention-Level Typing in FIGER}
To evaluate the efficacy of our methods we first compare against the current state-of-art models of \citet{shimaoka-EtAl:2017:EACLlong}. The most widely used type system for fine-grained entity typing is FIGER which consists of 113 types organized in a 2 level hierarchy. For training, we use the publicly available W2M data \citep{DBLP:conf/kdd/RenHQVJH16} and optimize the mention typing loss function defined in Section-\ref{sec:mention_typing} with the additional hierarchical loss where specified. For evaluation, we use the manually annotated FIGER (GOLD) data by \citet{ling2012fine}. See Appendix \ref{sec:model_implementation} and \ref{sec:figer_details} for specific implementation details.

\subsubsection{Results}
In Table \ref{tab:figer} we see that our base CNN models (CNN and CNN+Complex) match LSTM models of \citet{shimaoka-EtAl:2017:EACLlong} and \citet{ gupta2017EMNLP2017}, the previous state-of-the-art for models without hand-crafted features. When incorporating structure into our models, we gain 2.5 points of accuracy in our CNN+Complex model, matching the overall state of the art attentive LSTM that relied on handcrafted features from syntactic parses, topic models, and character n-grams. The structure can help our model predict lower frequency types which is a similar role played by hand-crafted features.

\begin{table}[t!]
\centering{
\small
\begin{tabular}{|l|rrr|}
\hline 
\multicolumn{1}{|c|}{Model} & Acc & Macro F1 & Micro F1 \\                           
\hline \hline
\citet{ling2012fine} & 47.4 & 69.2 & 65.5 \\
\citet{shimaoka-EtAl:2017:EACLlong} $\dagger$ & 55.6 & 75.1 & 71.7 \\
\citet{gupta2017EMNLP2017}$\dagger$ & 57.7 & 72.8 & 72.1 \\
\citet{shimaoka-EtAl:2017:EACLlong}$\ddagger$ & \bf{59.6} & \bf{78.9} & \bf{75.3} \\
\hline
CNN 		                      &  57.0   & 75.0  &   72.2   \\
+ hierarchy                       &  58.4  & 76.3  &  73.6    \\
CNN+Complex 		              &  57.2   & 75.3  &   72.9   \\
+ hierarchy                       &  \bf{59.7}   & \bf{78.3}  & \bf{75.4}   \\
\hline
\end{tabular}
\caption{Accuracy and Macro/Micro F1 on FIGER (GOLD). $\dagger$ is an LSTM model. $\ddagger$ is an attentive LSTM along with additional hand crafted features.\label{tab:figer}}
}
\vspace{-.15cm}
\end{table}

\subsection{Entity-Level Typing in TypeNet}
Next we evaluate our models on entity-level typing in TypeNet using Wikipedia. For each entity, we follow the procedure outlined in Section \ref{sec:entity_level_typing}. We predict labels for each instance in the entity's bag and aggregate them into entity-level predictions using $\LogSumExp$ pooling. Each type is assigned a predicted score by the model. We then rank these scores and calculate average precision for each of the types in the test set, and use these scores to calculate mean average precision (MAP). We evaluate using MAP instead of accuracy which is standard in large knowledge base link prediction tasks \cite{verga-neelakantan-mccallum:2017:EACLlong,trouillon2016complex}. These scores are calculated only over Freebase types, which tend to be lower in the hierarchy. This is to avoid artificial score inflation caused by trivial predictions such as `entity.' See Appendix \ref{sec:wiki_details} for more implementation details.

\subsubsection{Results}
Table \ref{typing-scores} shows the results for entity level typing on our Wikipedia TypeNet dataset. We see that both the basic CNN and the CNN+Complex models perform similarly with the CNN+Complex model doing slightly better on the full data regime. We also see that both models get an improvement when adding an explicit hierarchy loss, even before adding in the transitive closure. The transitive closure itself gives an additional increase in performance to both models. In both of these cases, the basic CNN model improves by a greater amount than CNN+Complex. This could be a result of the complex embeddings being more difficult to optimize and therefore more susceptible to variations in hyperparameters. When adding in both the transitive closure and the explicit hierarchy loss, the performance improves further. We observe similar trends when training our models in a lower data regime with \texttildelow150,000 examples, or about 5\% of the total data.

In all cases, we note that the baseline models that do not incorporate any hierarchical information (neither the transitive closure nor the hierarchy loss) perform \texttildelow9 MAP worse, demonstrating the benefits of incorporating structure information.

\begin{table}[t!]
\centering{
\small
\begin{tabular}{|l|rr|}
\hline 
Model & Low Data & Full Data   \\ 
\hline \hline
CNN 		                      &  51.72   	& 68.15         \\
+ hierarchy                       &  54.82   	& 75.56   		\\
+ transitive  					  &  57.68   	& 77.21    \\
+ hierarchy + transitive   		  &  58.74  & \textbf{78.59}   \\ 
\hline
CNN+Complex 		              &  50.51   	& 69.83         \\
+ hierarchy                       &  55.30    	& 72.86  	\\
+ transitive  					  &  53.71    	& 72.18      \\
+ hierarchy + transitive   		  &  \textbf{58.81} & 77.21  \\ \hline
\end{tabular}
\caption{MAP of entity-level typing in Wikipedia data using TypeNet. The second column shows results using 5\% of the total data. The last column shows results using the full set of 344,246 entities.}
\label{typing-scores}
}
\end{table}

\subsection{MedMentions Entity Linking with UMLS}

\begin{table}[t!]
	\centering
	{		\small		
		\begin{tabular}{|l|ll|}
			\hline 
			Model & original      & normalized             \\ \hline \hline
			mention tfidf    & \multicolumn{1}{r}{61.09}  & \multicolumn{1}{r|}{74.66}   \\ \hline
			CNN       & \multicolumn{1}{r}{67.42}   & \multicolumn{1}{r|}{82.40}           \\
            + hierarchy  & \multicolumn{1}{r}{67.73}   & \multicolumn{1}{r|}{82.77}     \\ \hline
			CNN+Complex    & \multicolumn{1}{r}{67.23}   & \multicolumn{1}{r|}{82.17}          \\ 
            + hierarchy & \multicolumn{1}{r}{\textbf{68.34}} & \multicolumn{1}{r|}{\textbf{83.52}} \\
            \hline
		\end{tabular}
		\caption{Accuracy on entity linking in MedMentions. Maximum recall is 81.82\% because we use an imperfect alias table to generate candidates. Normalized scores consider only mentions which contain the gold entity in the candidate set. Mention tfidf is $csim$ from Section \ref{sec:model_linking}.}
		\label{linking-scores}
	}
\end{table}

\begin{table*}[t!]
\centering
{
\small
\begin{tabular}{p{1.0\textwidth}} \hline
Tips and Pitfalls in \textbf{Direct Ligation} of Large Spontaneous Splenorenal Shunt during Liver Transplantation Patients with large spontaneous splenorenal shunt $\ldots$ \\
\textbf{baseline:} Direct [Direct $\rightarrow$ General Modifier $\rightarrow$ Qualifier $\rightarrow$ Property or Attribute] \\
\textbf{+hierarchy:} Ligature (correct) [Ligature $\rightarrow$ Surgical Procedures $\rightarrow$ medical treatment approach ] \\ 
 \hline
A novel approach for selective chemical functionalization and localized assembly of one-dimensional \textbf{nanostructures}. \\
\textbf{baseline:} Structure [Structure $\rightarrow$ order or structure $\rightarrow$ general epistemology] \\ 
\textbf{+hierarchy:} Nanomaterials (correct) [Nanomaterials $\rightarrow$  Nanoparticle Complex $\rightarrow$ Drug or Chemical by Structure] \\ 
 \hline
Gcn5 is recruited onto the \textbf{il-2} promoter by interacting with the NFAT in T cells upon TCR stimulation . \\
\textbf{baseline:} Interleukin-27 [Interleukin-27 $\rightarrow$ IL2 $\rightarrow$ Interleukin Gene] \\
\textbf{+hierarchy:} IL2 Gene (correct) [IL2 Gene $\rightarrow$ Interleukin Gene] \\
 \hline
\end{tabular}
\caption{Example predictions from MedMentions. Each example shows the sentence with entity mention span in bold. \textbf{Baseline}, shows the predicted entity and its ancestors of a model not incorporating structure. Finally, \textbf{+hierarchy} shows the prediction and ancestors for a model which explicitly incorporates the hierarchical structure information. \label{qual-linking}}
}
\end{table*}

In addition to entity typing, we evaluate our model's performance on an entity linking task using MedMentions, our new PubMed / UMLS dataset described in Section \ref{sec:pubmed_umls}.

\subsubsection{Results}
Table \ref{linking-scores} shows results for baselines and our proposed variant with additional hierarchical loss. None of these models incorporate transitive closure information, due to difficulty incorporating it in our candidate generation, which we leave to future work. The \textit{Normalized} metric considers performance only on mentions with an alias table hit; all models have 0 accuracy for mentions otherwise. We also report the overall score for comparison in future work with improved candidate generation. We see that incorporating structure information results in a 1.1\% reduction in absolute error, corresponding to a \texttildelow6\% reduction in relative error on this large-scale dataset.

Table \ref{qual-linking} shows qualitative predictions for models with and without hierarchy information incorporated. Each example contains the sentence (with target entity in bold), predictions for the baseline and hierarchy aware models, and the ancestors of the predicted entity. In the first and second example, the baseline model becomes extremely dependent on TFIDF string similarities when the gold candidate is rare ($\leq$ 10 occurrences). This shows that modeling the structure of the entity hierarchy helps the model disambiguate rare entities. In the third example, structure helps the model understand the hierarchical nature of the labels and prevents it from predicting an entity that is overly specific (e.g predicting Interleukin-27 rather than the correct and more general entity IL2 Gene). 

Note that, in contrast with the previous tasks, the complex hierarchical loss provides a significant boost, while the real-valued bilinear model does not. A possible explanation is that UMLS is a far larger/deeper ontology than even TypeNet, and the additional ability of complex embeddings to model intricate graph structure is key to realizing gains from hierarchical modeling. 

%% file: related_work.tex
\section{Related Work}

By directly linking a large set of mentions and typing a large set of entities with respect to a new ontology and corpus, and our incorporation of structural learning between the many entities and types in our ontologies of interest, our work draws on many different but complementary threads of research in information extraction, knowledge base population, and completion. 

Our structural, hierarchy-aware loss between types and entities draws on research in Knowledge Base Inference such as \citet{jain2018mitigating}, \citet{trouillon2016complex} and \citet{DBLP:conf/icml/NickelTK11}. Combining KB completion with hierarchical structure in knowledge bases has been explored in \cite{dalvi2015automatic, xie2016representation}. Recently, \citet{DBLP:journals/corr/abs-1709-01062} proposed a hierarchical loss for text classification.

Linking mentions to a flat set of entities, often in Freebase or Wikipedia, is a long-standing task in NLP \cite{bunescu2006using, cucerzan2007large, durrett2014joint, francis2016capturing}. Typing of mentions at varying levels of granularity, from CoNLL-style named entity recognition \cite{tjong2003introduction}, to the more fine-grained recent approaches \cite{ling2012fine,DBLP:journals/corr/GillickLGKH14, shimaoka-EtAl:2017:EACLlong}, is also related to our task. A few prior attempts to incorporate a very shallow hierarchy into fine-grained entity typing have not lead to significant or consistent improvements \cite{DBLP:journals/corr/GillickLGKH14,shimaoka-EtAl:2017:EACLlong}.

The knowledge base Yago \citep{DBLP:journals/ws/SuchanekKW08} includes integration with WordNet and type hierarchies have been derived from its type system \citep{yosef2012hyena}. \citet{delcorro-EtAl:2015:EMNLP} use manually crafted rules and patterns (Hearst patterns \cite{hearst1992automatic}, appositives, etc) to automatically match entity types to Wordnet synsets. 

Recent work has moved towards unifying these two highly related tasks by improving entity linking by simultaneously learning a fine grained entity type predictor \cite{gupta2017EMNLP2017}. Learning hierarchical structures or transitive relations between concepts has been the subject of much recent work \cite{vilnis2014word,vendrov2015order,nickel2017poincar}

We draw inspiration from all of this prior work, and contribute datasets and models to address previous challenges in jointly modeling the structure of large-scale hierarchical ontologies and mapping textual mentions into an extremely fine-grained space of entities and types.

%% file: appendix.tex
\section{Supplementary Materials}

\subsection{TypeNet Construction}
\begin{table}[!htb]{
\centering
\small
\begin{tabular}{|l|}
\hline
\textit{Freebase type}: musical\_chord                                                                             \\ \hline
\textit{Example entities}: psalms\_chord,  power\_chord \\\ \ \ \ \ \ \ \ \ \ \ \ \ \ \ \ \ \ \ \ \ \ \ \ \ \ \ \ \ \ harmonic\_seventh\_chord \\ 
\hline 
\hline
chord.n.01: a straight line connecting two points on a curve                                              \\ 
\hline
\textbf{chord.n.02}: a combination of three or more \\notes that blend harmoniously when sounded together   \\ 
\hline
musical.n.01: a play or film whose action and dialogue is\\ interspersed with singing and dancing           \\ 
\hline
\end{tabular}
\caption{Example given to TypeNet annotators. Here, the Freebase type to be linked is musical\_chord. This type is annotated in Freebase belonging to the entities psalms\_chord, harmonic\_seventh\_chord, and power\_chord. Below the list of example entities are candidate WordNet synsets obtained by substring matching between the Freebase type and all WordNet synsets. The correctly aligned synset is chord.n.02 shown in bold. \label{table:annotate}}
}
\end{table}

\subsection{Model Implementation Details \label{sec:model_implementation}} 
For all of our experiments, we use pretrained 300 dimensional word vectors from \citet{DBLP:conf/emnlp/PenningtonSM14}. These embeddings are fixed during training. The type vectors and entity vectors are all 300 dimensional vectors initialized using Glorot initialization \cite{DBLP:journals/jmlr/GlorotB10}. The number of negative links for hierarchical training $n \in$ \{16, 32, 64, 128, 256\}. 

For regularization, we use dropout \cite{DBLP:journals/jmlr/SrivastavaHKSS14} with p $\in$ \{0.5, 0.75, 0.8\} on the sentence encoder output and L2 regularize all learned parameters with $\lambda \in $ \{1e-5, 5e-5, 1e-4\}. All our parameters are optimized using Adam \citep{DBLP:journals/corr/KingmaB14} with a learning rate of 0.001. We tune our hyper-parameters via grid search and early stopping on the development set.  

\subsection{FIGER Implementation Details \label{sec:figer_details}}
To train our models, we use the mention typing loss function defined in Section-\ref{tab:figer}. For models with structure training, we additionally add in the hierarchical loss, along with a weight that is obtained by tuning on the dev set. We follow the same inference time procedure as \citet{shimaoka-EtAl:2017:EACLlong} For each mention, we first assign the type with the largest probability according to the logits, and then assign additional types based on the condition that their corresponding probability be greater than 0.5.

\subsection{Wikipedia Data and Implementation Details \label{sec:wiki_details}}
At train time, each training example randomly samples an entity bag of 10 mentions. At test time we classify bags of 20 mentions of an entity. The dataset contains a total of 344,246 entities mapped to the 1081 Freebase types from TypeNet. We consider all sentences in Wikipedia between 10 and 50 tokens long. Tokenization and sentence splitting was performed using NLTK \cite{loper2002nltk}. From these sentences, we considered all entities annotated with a cross-link in Wikipedia that we could link to Freebase and assign types in TypeNet. We then split the data by entities into a 90-5-5 train, dev, test split. 

\subsection{UMLS Implementation details \label{sec:umls_details}}
We pre-process each string by lowercasing and removing stop words. We consider ngrams from size 1 to 5 and keep the top 100,000 features and the final vectors are L2 normalized. For each mention,  In our experiments we consider the top 100 most similar entities as the candidate set.
\subsubsection{Candidate Generation Details \label{sec:candidate_details}}
Each mention and each canonical entity string in UMLS are mapped to TFIDF character ngram vectors. We pre-process each string by lowercasing and removing stop words. We consider ngrams from size 1 to 5 and keep the top 100,000 features and the final vectors are L2 normalized. For each mention, we calculate the cosine similarity, $\csim$, between the mention string and each canonical entity string. In our experiments we consider the top 100 most similar entities as the candidate set.